\documentclass[11pt,a4paper]{article}
\usepackage[nohyperref]{emnlp2018}
\usepackage{times}
\usepackage{latexsym}

\usepackage{url}

\usepackage{graphicx}
\usepackage{amsmath}
\usepackage{amsfonts}
\usepackage{booktabs}
\usepackage{multirow}
\usepackage{tikz}
\usepackage{pgfplots}
\usepackage{pgfplotstable}
\usepackage{lipsum}
\usepackage{enumitem}

\graphicspath{{./figures/}}
\pgfplotsset{compat=1.3}
\pgfplotstableset{col sep=comma}

\aclfinalcopy 

\newcommand\smalltalk{\textsc{Small-Talk}}
\newcommand\taskoriented{\textsc{Task-Oriented}}

\title{Empirical Evaluation of Character-Based Model on Neural Named-Entity
  Recognition in Indonesian Conversational Texts}

\author{Kemal Kurniawan \\
  Kata Research Team \\
  Kata.ai \\
  Jakarta, Indonesia \\
  {\tt kemal@kata.ai} \\\And
  Samuel Louvan \\
  Fondazione Bruno Kessler \\
  University of Trento \\
  Trento, Italy \\
  {\tt slouvan@fbk.eu} \\}

\date{}

\begin{document}
\maketitle
\begin{abstract}
  Despite the long history of named-entity recognition (NER) task in the natural
  language processing community, previous work rarely studied the task on
  conversational texts. Such texts are challenging because they contain a lot of
  word variations which increase the number of out-of-vocabulary (OOV) words.
  The high number of OOV words poses a difficulty for word-based neural models.
  Meanwhile, there is plenty of evidence to the effectiveness of
  character-based neural models in mitigating this OOV problem. We report an
  empirical evaluation of neural sequence labeling models with character
  embedding to tackle NER task in Indonesian conversational texts. Our
  experiments show that (1) character models outperform word embedding-only
  models by up to 4 $F_1$ points, (2) character models perform better in OOV
  cases with an improvement of as high as 15 $F_1$ points, and (3) character
  models are robust against a very high OOV rate.
\end{abstract}

\section{Introduction}

Critical to a conversational agent is the ability to recognize named entities.
For example, in a flight booking application, to book a ticket, the agent needs
information about the passenger's name, origin, and destination. While
named-entity recognition (NER) task has a long-standing history in the natural
language processing community, most of the studies have been focused on
recognizing entities in well-formed data, such as news articles or biomedical
texts. Hence, little is known about the suitability of the available named-entity
recognizers for conversational texts. In this work, we tried to shed some light
on this direction by evaluating neural sequence labeling models on NER task in
Indonesian conversational texts.

Unlike standard NLP corpora, conversational texts are typically noisy and
informal. For example, in Indonesian, the word \textit{aku} (``I'') can be
written as: \textit{aq}, \textit{akuw}, \textit{akuh}, \textit{q}. People also
tend to use non-standard words to represent named entities. This creative use of
language results in numerous word variations which may increase the number
out-of-vocabulary (OOV) words~\cite{baldwin2013}.

The most common approach to handle the OOV problem is by representing each OOV word
with a single vector representation (embedding). However, this treatment is not optimal
because it ignores the fact that words can share similar morphemes which can be
exploited to estimate the OOV word embedding better. Meanwhile, word
representation models based on subword units, such as characters or word
segments, have been shown to perform well in many NLP tasks such as POS
tagging~\cite{dossantos2014,ling2015}, language
modeling~\cite{ling2015,kim2016,vania2017}, machine
translation~\cite{vylomova2016,lee2016,sennrich2016}, dependency
parsing~\cite{ballesteros2015}, and sequence labeling~\cite{rei2016,lample2016}.
These representations are effective because they can represent OOV words better
by leveraging the orthographic similarity among words.

As for Indonesian NER, the earliest work was done by~\newcite{budi2005} which
relied on a rule-based approach. More recent research mainly used machine
learning methods such as conditional random
fields (CRF)~\cite{luthfi2014,leonandya2015,taufik2016} and support vector
machines~\cite{suwarningsih2014,aryoyudanta2016}. The most commonly used
datasets are news articles~\cite{budi2005}, Wikipedia/DBPedia
articles~\cite{luthfi2014,leonandya2015,aryoyudanta2016}, medical
texts~\cite{suwarningsih2014}, and Twitter data~\cite{taufik2016}.
To the best of our knowledge, there has been no work that used neural
networks for Indonesian NER nor NER for Indonesian conversational texts.

In this paper, we report the ability of a neural network-based approach for
Indonesian NER in conversational data. We employed the neural sequence labeling
model of~\cite{rei2016} and experimented with two word representation models:
word-level and character-level. We evaluated all models on relatively large,
manually annotated Indonesian conversational texts. We aim to address the
following questions:
\begin{enumerate}[noitemsep,topsep=0pt,label={\arabic*)}]
  \item How do the character models perform compared to word embedding-only
    models on NER in Indonesian conversational texts?
  \item How much can we gain in terms of performance from using the character
    models on OOV cases?
  \item How robust (in terms of performance) are the character models on
    different levels of OOV rates?
\end{enumerate}
Our experiments show that (1) the character models perform really well compared
to word embedding-only with an improvement up to 4 $F_1$ points, (2) we can gain
as high as 15 $F_1$ points on OOV cases by employing character models, and (3)
the character models are highly robust against OOV rate as there is no
noticeable performance degradation even when the OOV rate approaches 100\%.

\section{Methodology}

\begin{table*}[t]\small
  \centering
  \begin{tabular}{@{}ll@{}}
    \toprule
    Dataset & Example \\
    \midrule
    \multirow{4}{*}{\smalltalk{}}
            & sama2 sumatera barat, tapi gue di Pariaman bkn Payakumbuh \\
            & ``also in west sumatera, but I am in pariaman not payakumbuh'' \\
            & rere jem rere, bukan riri. Riri itu siapa deeeh \\
            & ``(it's) rere jem rere, not riri. who's riri?'' \\
    \midrule[0.5\lightrulewidth]
    \multirow{4}{*}{\taskoriented{}}
            & Bioskop di lippo mall jogja brapa bos? \\
            & ``how much does the movies at lippo mall jogja cost?'' \\
            & Tolong cariin nomor telepon martabak pecenongan kelapa gading, sama tutup jam brp \\
            & ``please find me the phone number for martabak pecenongan kelapa gading, and what time it closes'' \\
    \bottomrule
  \end{tabular}
  \caption{Example texts from each dataset. \smalltalk{} contains small
    talk conversations, while \taskoriented{} contains task-oriented
    imperative texts such as flight booking or food delivery. English translations
    are enclosed in quotes.}\label{tbl:samples}
\end{table*}

\begin{table}[t]\scriptsize
  \centering
  \begin{tabular}{@{}ll@{}}
    \toprule
    Entity       & Example                                              \\
    \midrule
    DATETIME     & 17 agustus 1999, 15februari2001, 180900              \\
    EMAIL        & dianu\#\#\#\#\#@yahoo.co.id, b.s\#\#\#\#\#@gmail.com \\
    GENDER       & pria, laki, wanita, cewek                            \\
    LOCATION/LOC & salatiga, Perumahan Griya Mawar Sembada Indah        \\
    PERSON/PER   & Yusan Darmaga, Natsumi Aida, valentino rossi         \\
    PHONE        & 085599837\#\#\#, 0819.90.837.\#\#\#                  \\
    \bottomrule
  \end{tabular}
  \caption{Some examples of each entity. Some parts are replaced with \#\#\# for
    privacy reasons.}\label{tbl:ex-ents}
\end{table}

\begin{table*}[t]\small
  \centering
  \begin{tabular}{@{}rrrrrrrrrr@{}}
    \toprule
    \multicolumn{6}{c}{\smalltalk{}} & \multicolumn{4}{c}{\taskoriented{}} \\
    \cmidrule(r){1-6} \cmidrule(l){7-10}
    DATETIME & EMAIL & GENDER & LOCATION & PERSON & PHONE & EMAIL & LOC   & PER   & PHONE \\
    \midrule
    90       & 35    & 390    & 4352     & 3958   & 83    & 1707  & 55614 & 40624 & 3186  \\
    \bottomrule
  \end{tabular}
  \caption{Number of entities in both datasets.}\label{tbl:tag-dist}
\end{table*}

\begin{table}[t]\small
  \centering
  \begin{tabular}{@{}llrr@{}}
    \toprule
                         &        & \smalltalk{} & \taskoriented{} \\
    \midrule
    \multirow{3}{*}{$L$} & mean   & 3.63         & 14.84           \\
                         & median & 3.00         & 12.00           \\
                         & std    & 2.68         & 11.50           \\
    \midrule[0.5\lightrulewidth]
    \multirow{3}{*}{$N$} & train  & 10 044       & 51 120          \\
                         & dev    & 3 228        & 14 354          \\
                         & test   & 3 120        & 7 097           \\
    \midrule[0.5\lightrulewidth]
    \multirow{2}{*}{$O$} & dev    & 57.59        & 41.39           \\
                         & test   & 57.79        & 32.17           \\
    \bottomrule
  \end{tabular}
  \caption{Sentence length ($L$), number of sentences ($N$), and OOV rate ($O$)
    in each dataset. Sentence length is measured by the number of words. OOV
    rate is the proportion of word types that do not occur in the training
    split.}\label{tbl:datastats}
\end{table}

We used our own manually annotated datasets collected from users using our
chatbot service. There are two datasets: \smalltalk{} and \taskoriented{}.
\smalltalk{} contains 16K conversational messages from our users having small
talk with our chatbot, Jemma.\footnote{Available at LINE messaging as @jemma.}
\taskoriented{} contains 72K task-oriented imperative messages such as flight
booking, food delivery, and so forth obtained from YesBoss
service.\footnote{YesBoss is our hybrid virtual assistant service.} Thus,
\taskoriented{} usually has longer texts and more precise entities (e.g.,
locations) compared to \smalltalk{}. Table~\ref{tbl:samples} shows some example
sentences for each dataset. A total of 13 human annotators annotated the two
datasets. Unfortunately, we cannot publish the datasets because of proprietary
reasons.

\smalltalk{} has 6 entities: DATETIME, EMAIL, GENDER, LOCATION, PERSON, and PHONE.
\taskoriented{} has 4 entities: EMAIL, LOC, PER, and PHONE. The two datasets
have different entity inventory because the two chatbot purposes are different.
In \smalltalk{}, we care about personal information such as date of birth,
email, or gender to offer personalized content. In \taskoriented{}, the tasks
usually can be performed by providing minimal personal information. Therefore,
some of the entities are not necessary. Table~\ref{tbl:ex-ents}
and~\ref{tbl:tag-dist} report some examples of each entity and the number of
entities in both datasets respectively. The datasets are tagged using BIO
tagging scheme and split into training, development, and testing set. The
complete dataset statistics, along with the OOV rate for each split, are shown
in Table~\ref{tbl:datastats}. We define OOV rate as the percentage of word types
that do not occur in the training set. As seen in the table, the OOV rate is
quite high, especially for \smalltalk{} with more than 50\% OOV rate.

As baselines, we used a simple model which memorizes the word-tag assignments on
the training data~\cite{nadeau2007} and a feature-based CRF~\cite{lafferty2001},
as it is a common model for Indonesian NER. We used almost identical features
as~\newcite{taufik2016} since they experimented on the Twitter dataset which we
regarded as the most similar to our conversational texts among other previous
work on Indonesian NER. Some features that we did not employ were POS tags,
lookup list, and non-standard word list as we did not have POS tags in our data
nor access to the lists~\newcite{taufik2016} used. For the CRF model, we used an
implementation provided
by~\newcite{okazaki2007}\footnote{http://www.chokkan.org/software/crfsuite/}.

Neural architectures for sequence labeling are pretty similar. They usually
employ a bidirectional LSTM~\cite{hochreiter1997} with CRF as the output layer,
and a CNN~\cite{ma2016} or LSTM~\cite{lample2016,rei2016} composes the character
embeddings. Also, we do not try to achieve state-of-the-art
results but only are interested whether neural sequence labeling models with
character embedding can handle the OOV problem well. Therefore, for the neural
models, we just picked the implementation provided
in~\cite{rei2016}.\footnote{https://github.com/marekrei/sequence-labeler}

In their implementation, all the LSTMs have only one layer.
Dropout~\cite{srivastava2014} is used as the regularizer but only applied to the
final word embedding as opposed to the LSTM outputs as proposed
by~\newcite{zaremba2015}. The loss function contains not only the log likelihood of
the training data and the similarity score but also a language modeling loss,
which is not mentioned in~\cite{rei2016} but discussed in the subsequent
work~\cite{rei2017}. Thus, their implementation essentially does multi-task
learning with sequence labeling as the primary task and language modeling as the
auxiliary task.

We used an almost identical setting to~\newcite{rei2016}: words are lowercased,
but characters are not, digits are replaced with zeros, singleton words in the
training set are converted into unknown tokens, word and character embedding
sizes are 300 and 50 respectively. The character embeddings were initialized
randomly and learned during training. LSTMs are set to have 200 hidden units,
the pre-output layer has an output size of 50, CRF layer is used as the output
layer, and early stopping is used with a patience of 7. Some differences are: we
did not use any pretrained word embedding, and we used Adam
optimization~\cite{kingma2014} with a learning rate of 0.001 and batch size of
16 to reduce GPU memory usage. We decided not to use any pretrained word
embedding because to the best of our knowledge, there is no off-the-shelf
Indonesian pretrained word embedding that is trained on conversational data. The
ones available are usually trained on Wikipedia articles (\texttt{fastText}) and
we believe it has a very small size of shared vocabulary with conversational
texts. We tuned the dropout rate on the development set via grid search, trying
multiples of 0.1. We evaluated all of our models using CoNLL evaluation:
micro-averaged $F_1$ score based on exact span matching.

\section{Results and discussion}

\subsection{Performance}

\begin{table}[t]\small
  \centering
  \begin{tabular}{@{}lrr@{}}
    \toprule
    Model           & \smalltalk{}   & \taskoriented{} \\
    \midrule
    \textsc{Memo}   & 38.03          & 46.35           \\
    \textsc{Crf}    & 75.50          & 73.25           \\
    \midrule[0.5\lightrulewidth]
    \textsc{Word}   & 80.96          & 79.35           \\
    \textsc{Concat} & 84.73          & \textbf{80.22}  \\
    \textsc{Attn}   & \textbf{84.97} & 79.71           \\
    \bottomrule
  \end{tabular}
  \caption{$F_1$ scores on the test set of each dataset. The scores are computed
    as in CoNLL evaluation. \textsc{Memo}: memorization baseline. \textsc{Crf}:
    CRF baseline. \textsc{Word}, \textsc{Concat}, \textsc{Attn}:
    \citeauthor{rei2016}'s word embedding-only, concatenation, and attention
    model respectively.}\label{tbl:f1}
\end{table}

\begin{figure*}[t]
  \centering
  \includegraphics[width=0.4\textwidth]{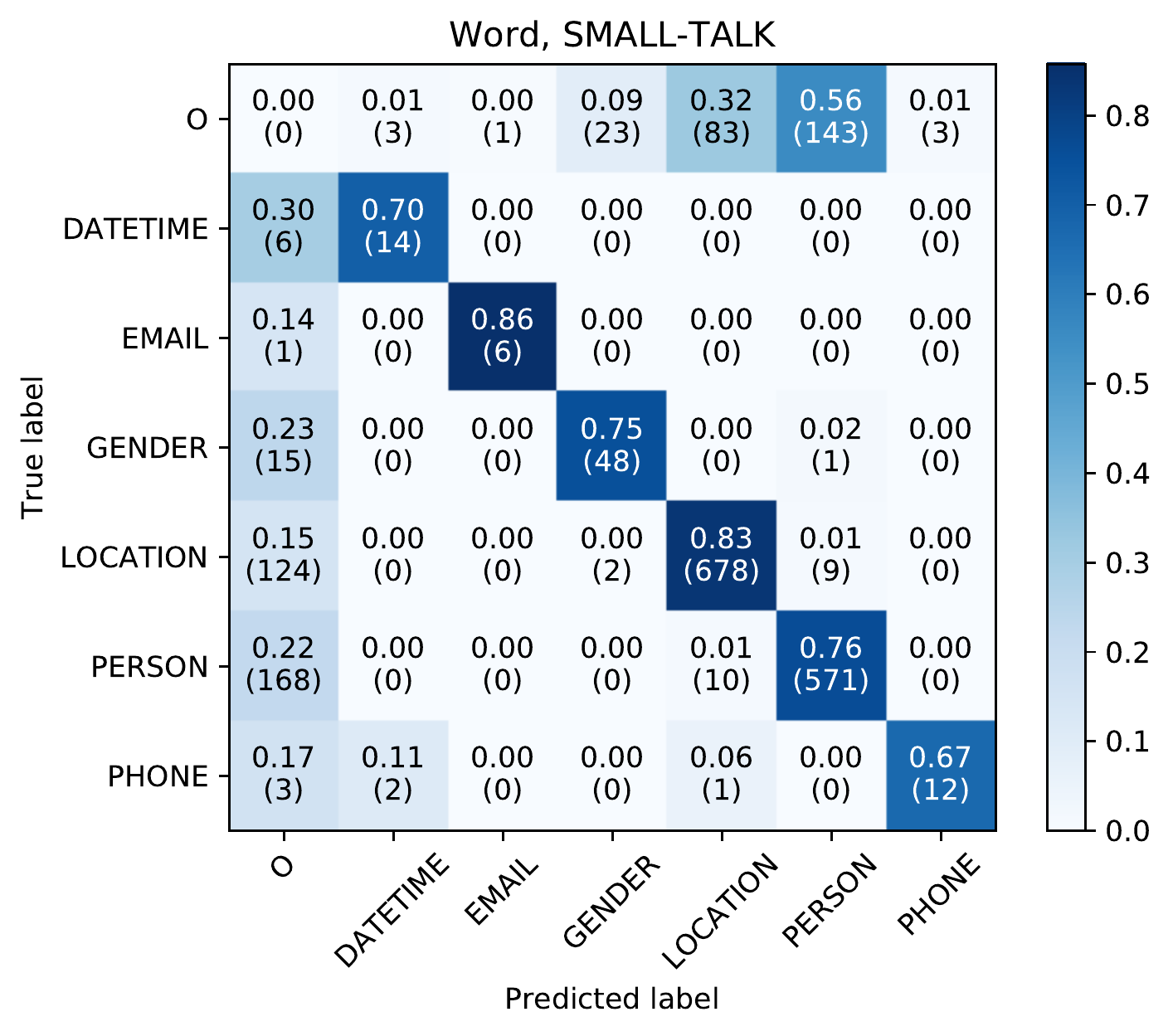}
  \includegraphics[width=0.4\textwidth]{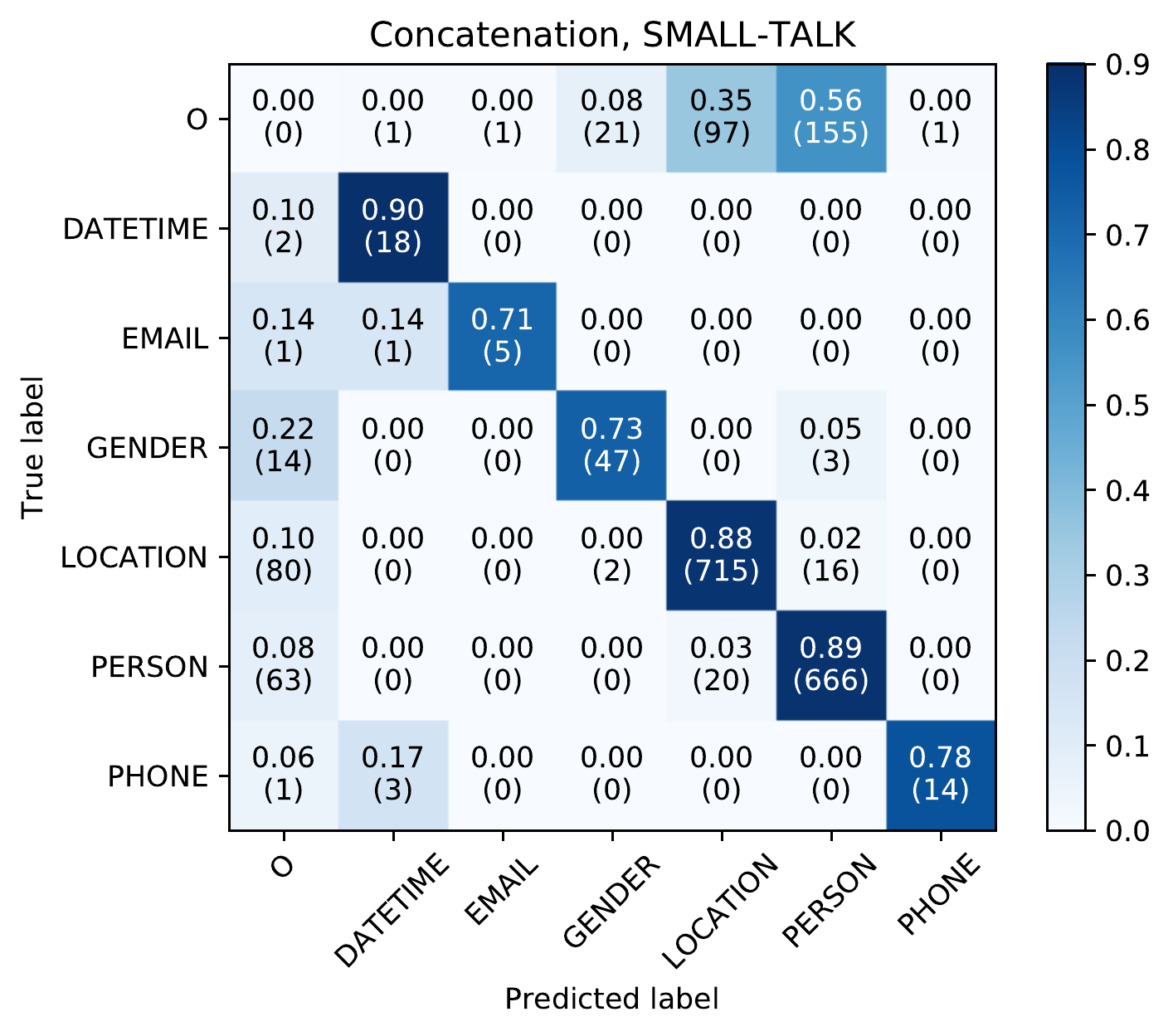}
  \includegraphics[width=0.4\textwidth]{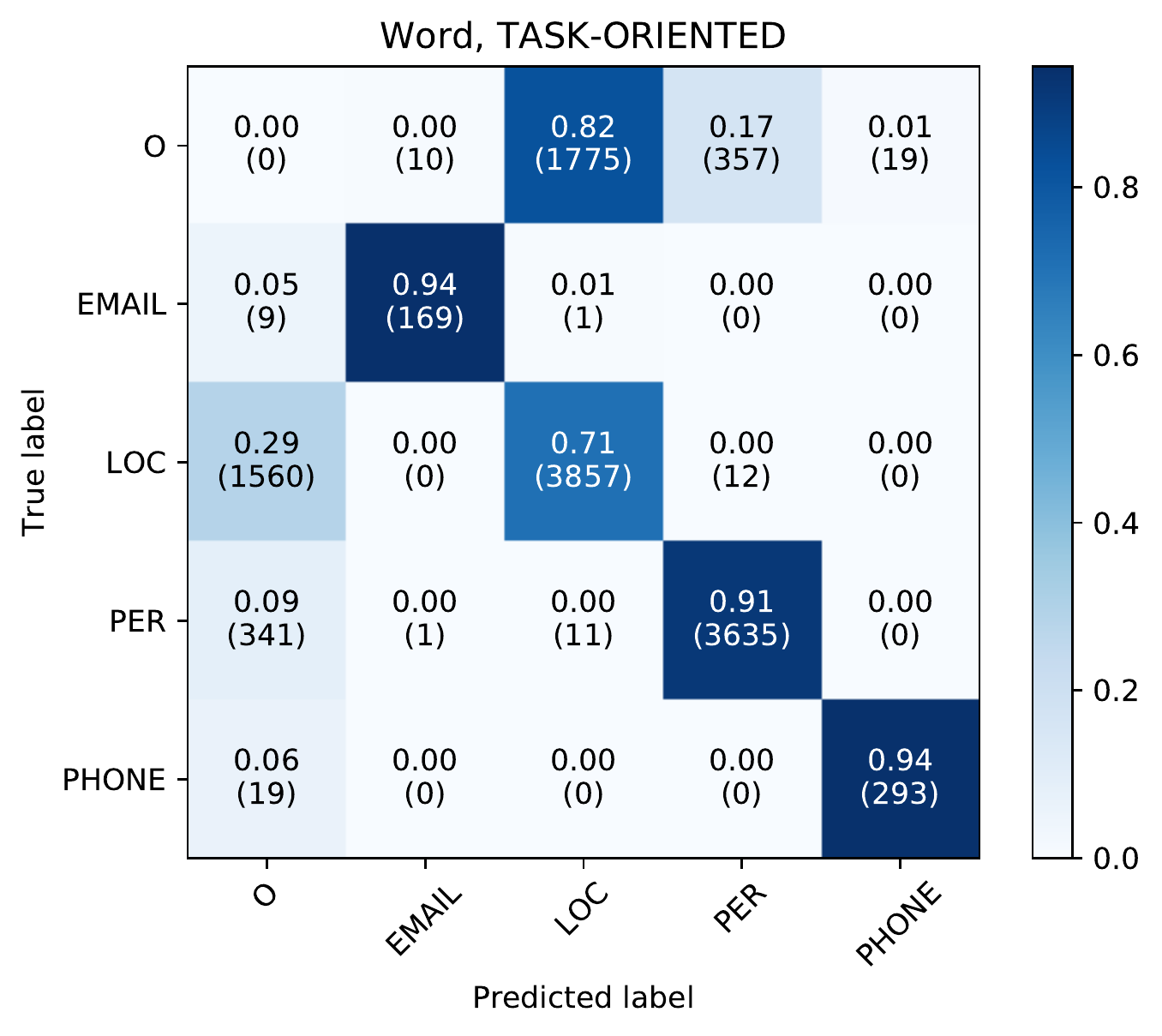}
  \includegraphics[width=0.4\textwidth]{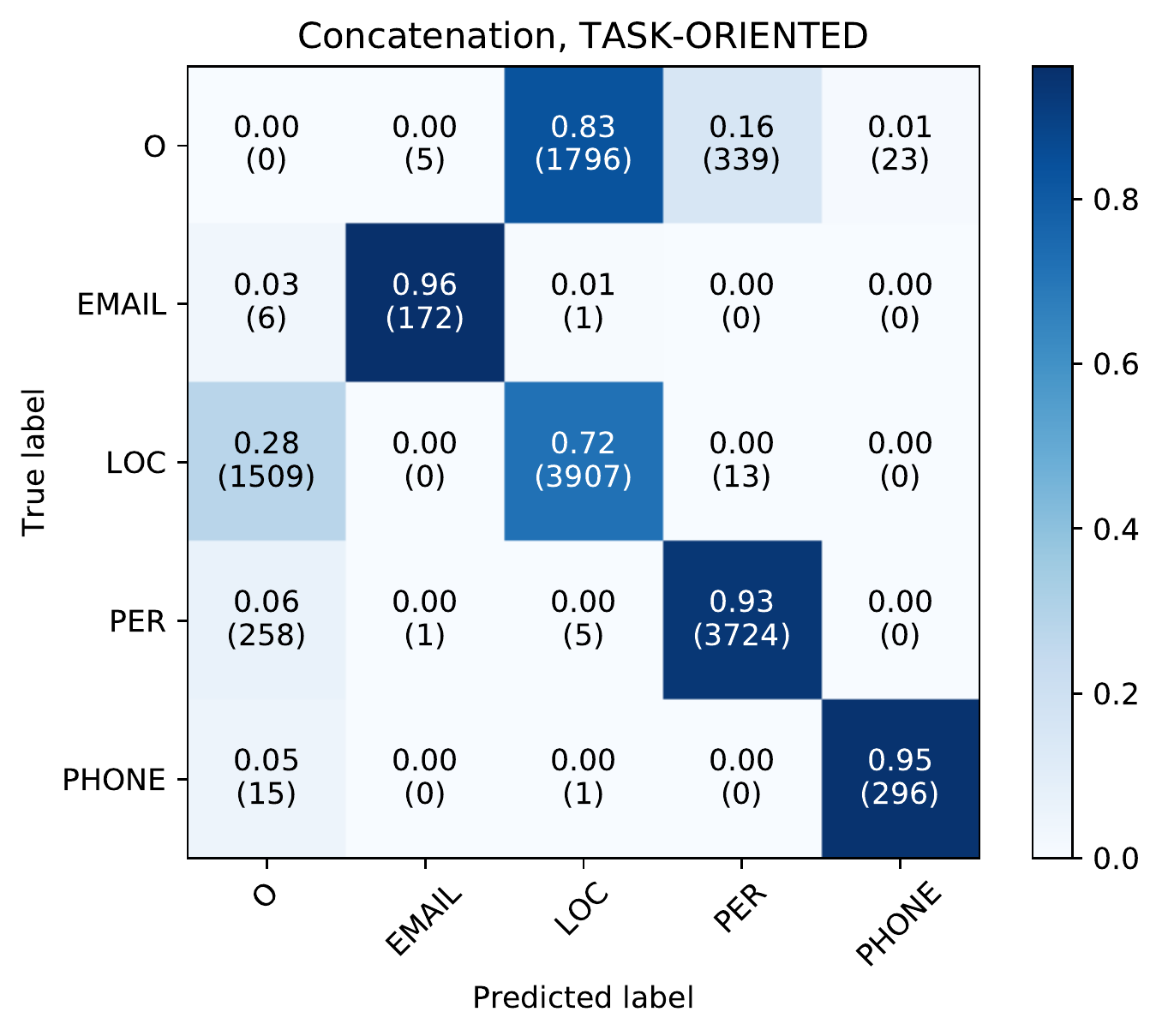}
  \caption{Confusion matrices of the word embedding-only and concatenation model
    on the test set of each dataset. Top row: \smalltalk{} dataset. Bottom
    row: \taskoriented{} dataset. Left column: word embedding-only model. Right column:
    concatenation model.}~\label{fig:confmat}
\end{figure*}

\begin{table}[t]\scriptsize
  \centering
  \begin{tabular}{@{}lllll@{}}
    \toprule
    token     & vocab        & gold           & word           & concat         \\
    \midrule
    Kantor    & kantor       & \texttt{B-LOC} & \texttt{B-LOC} & \texttt{B-LOC} \\
    PKPK      & \texttt{UNK} & \texttt{I-LOC} & \texttt{I-LOC} & \texttt{I-LOC} \\
    lt        & lt           & \texttt{I-LOC} & \texttt{B-LOC} & \texttt{B-LOC} \\
    .         & .            & \texttt{I-LOC} & \texttt{I-LOC} & \texttt{I-LOC} \\
    3         & 0            & \texttt{I-LOC} & \texttt{I-LOC} & \texttt{I-LOC} \\
    ,         & ,            & \texttt{O}     & \texttt{O}     & \texttt{O}     \\
    Gedung    & gedung       & \texttt{B-LOC} & \texttt{B-LOC} & \texttt{B-LOC} \\
    Fak       & \texttt{UNK} & \texttt{I-LOC} & \texttt{I-LOC} & \texttt{I-LOC} \\
    .         & .            & \texttt{I-LOC} & \texttt{O}     & \texttt{I-LOC} \\
    Psikologi & psikologi    & \texttt{I-LOC} & \texttt{B-LOC} & \texttt{I-LOC} \\
    UNAIR     & unair        & \texttt{I-LOC} & \texttt{I-LOC} & \texttt{I-LOC} \\
    Kampus    & kampus       & \texttt{I-LOC} & \texttt{B-LOC} & \texttt{B-LOC} \\
    B         & b            & \texttt{I-LOC} & \texttt{I-LOC} & \texttt{B-LOC} \\
    .         & .            & \texttt{O}     & \texttt{O}     & \texttt{O}     \\
    Jl        & jl           & \texttt{B-LOC} & \texttt{B-LOC} & \texttt{B-LOC} \\
    Airlangga & airlangga    & \texttt{I-LOC} & \texttt{I-LOC} & \texttt{I-LOC} \\
    no        & no           & \texttt{I-LOC} & \texttt{I-LOC} & \texttt{I-LOC} \\
    .         & .            & \texttt{I-LOC} & \texttt{I-LOC} & \texttt{I-LOC} \\
    4-6       & \texttt{UNK} & \texttt{I-LOC} & \texttt{I-LOC} & \texttt{I-LOC} \\
    Sby       & sby          & \texttt{I-LOC} & \texttt{B-LOC} & \texttt{B-LOC} \\
    \bottomrule
  \end{tabular}
  \caption{An example displaying how the word embedding-only (word) and
    concatenation (concat) models can partition a long location entity into its
    parts.}\label{tbl:yb-loc}
\end{table}

Table~\ref{tbl:f1} shows the overall $F_1$ score on the test
set of each dataset. We see that the neural network models beat both baseline
models significantly. We also see that the character models consistently
outperform the word embedding-only model, where the improvement can be as high
as 4 points on \smalltalk{}. An interesting observation is how the improvement
is much larger in \smalltalk{} than \taskoriented{}. We speculate that this is due
to the higher OOV rate \smalltalk{} has, as can be seen in
Table~\ref{tbl:datastats}.

To understand the character model better, we draw the confusion matrix of the
word embedding-only and the concatenation model for each dataset in
Figure~\ref{fig:confmat}. We chose only the concatenation model because both
character models are better than the word embedding-only, so we just picked the
simplest one.

\textbf{\smalltalk{}}. Both word embedding-only and
concatenation model seem to hallucinate PERSON and LOCATION often. This
observation is indicated by the high false positive rate of those entities,
where 56\% of non-entities are recognized as PERSON, and about 30\% of
non-entities are recognized as LOCATION. Both models appear to confuse PHONE as
DATETIME as marked by 11\% and 17\% misclassification rate of the models
respectively.

The two models also have some differences. The word embedding-only model has
higher false negative than the concatenation model. DATETIME has the highest false
negative, where the word embedding-only model incorrectly classified 30\% of
true entities as non-entity. Turning to the concatenation model, we see how the
false negative decreases for almost all entities. DATETIME has the most significant
drop of 20\% (down from 30\% to 10\%), followed by PERSON, PHONE, LOCATION,
and GENDER.

\textbf{\taskoriented{}}. The confusion matrices of
the two models are strikingly similar. The models seem to have a hard time
dealing with LOC because it often hallucinates the existence of LOC (as
indicated by the high false positive rate) and misses genuine LOC entities (as
shown by the high false negative rate). Upon closer look, we found that the two
models actually can recognize LOC well, but sometimes they partition it into its
parts while the gold annotation treats the entity as a single unit.
Table~\ref{tbl:yb-loc} shows an example of such case. A long location
like \textit{Kantor PKPK lt. 3} is partitioned by the models into \textit{Kantor
  PKPK} (office name) and \textit{lt. 3} (floor number). The models also
partition \textit{Jl Airlangga no. 4-6 Sby} into \textit{Jl Airlangga no. 4-6}
(street and building number) and \textit{Sby} (abbreviated city name). We think
that this partitioning behavior is reasonable because each part is indeed a
location.

There is also some amount of false positive on PER, signaling that the models
sometimes falsely recognize a non-entity as a person's name. The similarity of
the two confusion matrices appears to demonstrate that character embedding only
provides a small improvement on the \taskoriented{} dataset.

\subsection{Performance on OOV entities}

Next, we want to
understand better how much gain we can get from character models on OOV cases.
To answer this question, we ignored entities that do not have any OOV word on
the test set and re-evaluated the word embedding-only and concatenation models.
Table~\ref{tbl:f1-oov-span} shows the re-evaluated overall and
per-entity $F_1$ score on the test set of each dataset. We see how the
concatenation model consistently outperforms the word embedding-only model for
almost all entities on both datasets. On \smalltalk{} dataset, the overall $F_1$
score gap is as high as 15 points. It is also remarkable that the
concatenation model manages to achieve 40 $F_1$ points for GENDER on \smalltalk{}
while the word embedding-only cannot even recognize any GENDER.
Therefore, in general, this result corroborates our hypothesis that the
character model is indeed better at dealing with the OOV problem.

\begin{table*}[t]\small
  \centering
  \begin{tabular}{@{}lrr@{}}
    \toprule
    Entity   & word            & concat         \\
    \midrule
    DATETIME & 50.00           & \textbf{87.50} \\
    EMAIL    & \textbf{100.00} & 88.89          \\
    GENDER   & *0.00           & \textbf{40.00} \\
    LOCATION & 51.38           & \textbf{63.18} \\
    PERSON   & 68.36           & \textbf{80.14} \\
    PHONE    & 0.00            & \textbf{40.00} \\
    \cmidrule[0.5\lightrulewidth]{1-3}
    Overall  & 46.14           & \textbf{61.75} \\
    \bottomrule
  \end{tabular}
  \hspace{10mm}
  \begin{tabular}{@{}lrr@{}}
    \toprule
    Entity   & word            & concat         \\
    \midrule
    EMAIL    & 95.06           & \textbf{96.59} \\
    LOC      & 54.49           & \textbf{54.74} \\
    PER      & 73.22           & \textbf{82.55} \\
    PHONE    & *\textbf{0.00}  & \textbf{0.00}  \\
    \cmidrule[0.5\lightrulewidth]{1-3}
    Overall  & 50.05           & \textbf{54.54} \\
    \bottomrule
  \end{tabular}
  \caption{$F_1$ scores of word embedding-only (word) and concatenation
    (concat) model on the test set of \smalltalk{} (left) and \taskoriented{}
    (right) but \textbf{only} for entities containing at least one OOV word.
    Entries marked with an asterisk (*) indicate that the model does not
    recognize any entity at all.}\label{tbl:f1-oov-span}
\end{table*}

\subsection{Impact of OOV rate to model performance}

\begin{figure*}[!htbp]\small
  \centering
  \begin{tikzpicture}
    \begin{axis}[
      width=0.5\textwidth,
      title=\smalltalk{},
      xlabel={Threshold\\(OOV rate \%)},
      xlabel style={align=center},
      ylabel={$F_1$},
      ymin=0.1, ymax=0.9,
      ytick={0.2, 0.4, 0.6, 0.8},
      xmode=log,
      log basis x={2},
      xtick={1,2,4,8,16,32,64,128,256,512,1024},
      xticklabels={{$2^0$\\\tiny{(43.5)}}, {$2^1$\\\tiny{(57.8)}}, {$2^2$\\\tiny{(71.3)}}, {$2^3$\\\tiny{(83.0)}}, {$2^4$\\\tiny{(90.8)}}, {$2^5$\\\tiny{(95.3)}}, {$2^6$\\\tiny{(97.6)}}, {$2^7$\\\tiny{(98.8)}}, {$2^8$\\\tiny{(99.5)}}, {$2^9$\\\tiny{(99.8)}}, {$2^{10}$\\\tiny{(99.9)}}},
      xticklabel style={align=center},
      cycle list name=black white,
      grid=major,
      grid style=dashed,
    ]
    \pgfplotstableread{results/jemma-vary-oov-rate.dat}\datatable
    \addplot table[x=Threshold, y=noint] {\datatable};
    \addplot table[x=Threshold, y=concat] {\datatable};
    \addplot table[x=Threshold, y=attn] {\datatable};
    \end{axis}
  \end{tikzpicture}
  \hspace{1mm}
  \begin{tikzpicture}
    \begin{axis}[
      width=0.5\textwidth,
      title=\taskoriented{},
      xlabel={Threshold\\(OOV rate \%)},
      xlabel style={align=center},
      ylabel={$F_1$},
      ymin=0.1, ymax=0.9,
      ytick={0.2, 0.4, 0.6, 0.8},
      xmode=log,
      log basis x={4},
      xtick={1,4,16,64,256,1024,4096,16384},
      xticklabels={{$4^0$\\\tiny{(21.6)}}, {$4^1$\\\tiny{(44.9)}}, {$4^2$\\\tiny{(72.2)}}, {$4^3$\\\tiny{(89.8)}}, {$4^4$\\\tiny{(96.8)}}, {$4^5$\\\tiny{(99.1)}}, {$4^6$\\\tiny{(99.8)}}, {$4^7$\\\tiny{(100)}}},
      xticklabel style={align=center},
      cycle list name=black white,
      grid=major,
      grid style=dashed,
      legend columns=-1,
      legend to name=legend_name,
    ]
    \pgfplotstableread{results/yb-vary-oov-rate.dat}\datatable
    \addplot table[x=Threshold, y=noint] {\datatable};
    \addlegendentry{\textsc{Word}}
    \addplot table[x=Threshold, y=concat] {\datatable};
    \addlegendentry{\textsc{Concat}}
    \addplot table[x=Threshold, y=attn] {\datatable};
    \addlegendentry{\textsc{Attn}}
    \end{axis}
  \end{tikzpicture}
  \ref{legend_name}
  \caption{$F_1$ scores on the test set of each dataset with varying
    threshold. Words occurring fewer than this threshold in the training set are
    converted into the special token for OOV words. OOV rate increases as threshold
    does (from left to right). \textsc{Word}, \textsc{Concat}, and \textsc{Attn}
    refers to the word embedding-only, concatenation, and attention model
    respectively.}\label{fig:vary}
\end{figure*}
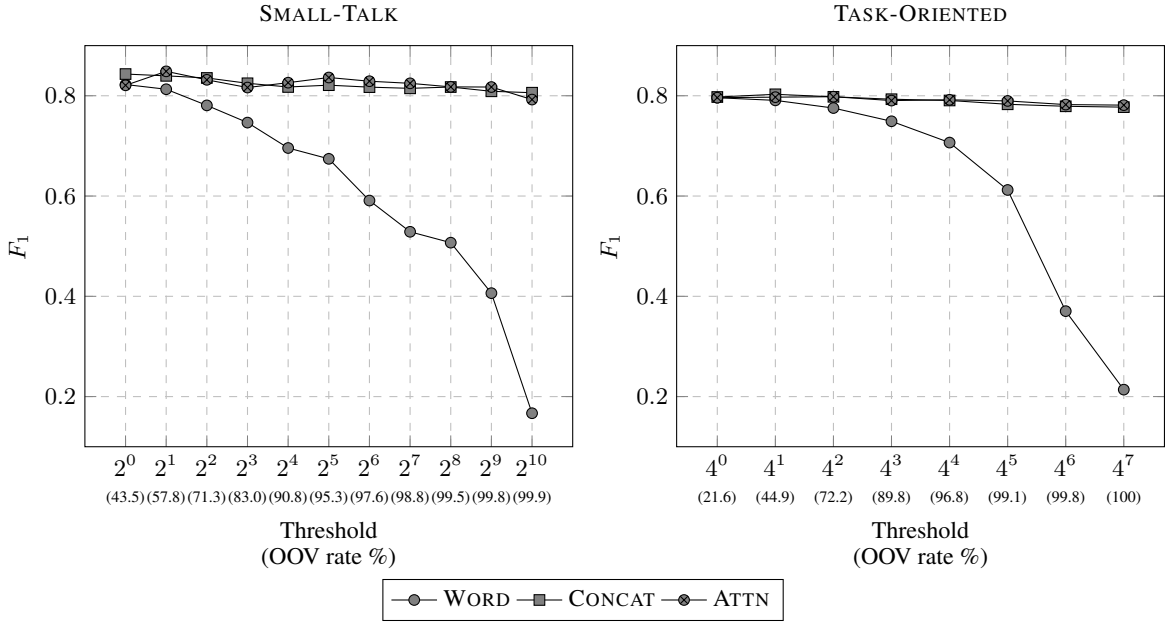

To better understand to what extent the character models can mitigate OOV
problem, we evaluated the performance of the models on different OOV rates. We
experimented by varying the OOV rate on each dataset and plot the result in
Figure~\ref{fig:vary}. Varying the OOV rate can be achieved by changing the
minimum frequency threshold for a word to be included in the vocabulary. Words
that occur fewer than this threshold in the training set are converted into the
special token for OOV words. Thus, increasing this threshold means increasing
the OOV rate and vice versa.

From Figure~\ref{fig:vary}, we see that across all datasets, the models which
employ character embedding, either by concatenation or attention, consistently
outperform the word embedding-only model at almost every threshold level. The
performance gap is even more pronounced when the OOV rate is high. Going from left
to right, as the OOV rate increases, the character models performance does not
seem to degrade much. Remarkably, this is true even when OOV rate is as high as 90\%,
even approaching 100\%, whereas the word embedding-only model already has
a significant drop in performance when the OOV rate is just around 70\%. This
finding confirms that character embedding is useful to mitigate the OOV
problem and robust against different OOV rates. We also observe that there seems
no perceptible difference between the concatenation and attention model.

\section{Conclusion and future work}

We reported an empirical evaluation of neural sequence labeling models
by~\newcite{rei2016} on NER in Indonesian conversational texts. The neural
models, even without character embedding, outperform the CRF baseline, which is
a typical model for Indonesian NER. The models employing character embedding
have an improvement up to 4 $F_1$ points compared to the word embedding-only
counterpart. We demonstrated that by using character embedding, we could gain
improvement as high as 15 $F_1$ points on entities having OOV words. Further
experiments on different OOV rates show that the character models are highly robust
against OOV words, as the performance does not seem to degrade even when the OOV
rate approaches 100\%.

While the character model by~\newcite{rei2016} has produced good results, it is
still quite slow because of the LSTM used for composing character embeddings.
Recent work on sequence labeling by~\newcite{reimers2017} showed that replacing
LSTM with CNN for composition has no significant performance drop but is faster
because unlike LSTM, CNN computation can be parallelized. Using character
trigrams as subword units can also be an avenue for future research, as their
effectiveness has been shown by~\newcite{vania2017}. Entities like PHONE and
EMAIL have quite clear patterns so it might be better to employ a
regex-based classifier to recognize such entities and let the neural network
models tag only person and location names.




\bibliographystyle{acl_natbib_nourl}
\bibliography{oov}

\end{document}